%% file: main.tex
\documentclass{article}

\PassOptionsToPackage{numbers,compress}{natbib}
 \usepackage[preprint]{neurips_2026}

\usepackage[utf8]{inputenc} 
\usepackage[T1]{fontenc}    
\usepackage{hyperref}       
\usepackage{url}            
\usepackage{booktabs}       
\usepackage{amsfonts}       
\usepackage{nicefrac}       
\usepackage{microtype}      
\usepackage{xcolor}         
\usepackage{graphicx}
\usepackage{subcaption}
\usepackage{multirow}
\usepackage{xcolor}
\usepackage{wrapfig}
\usepackage{algorithm}
\usepackage{algpseudocode}
\usepackage{xurl}
\usepackage{capt-of}
\usepackage{amsmath}

\title{Eliciting Self-Verification in Multimodal Reasoning Agents with Reinforcement Learning}

\author{%
  Vishwas Sathish\thanks{Equal contribution.} \hspace{0.10em} \thanks{Corresponding author. \protect\\    \hspace*{1.5em} Project website: \url{https://vishwassathish.github.io/projects/svrl/}} \\
  University of Washington \\
  Seattle, WA, USA \\
  \texttt{vsathish@cs.washington.edu} \\
  \And
  Viresh Ranjan\footnotemark[1] \\
  Amazon \\
  USA \\
  \texttt{vireshr@amazon.com} \\
  \And
  Xinliang Zhu \\
  Amazon \\
  USA \\
  \texttt{xlzhu@amazon.com} \\
  \AND
  Arnab Dhua \\
  Amazon \\
  USA \\
  \texttt{adhua@amazon.com} \\
  \And
  Douglas Gray \\
  Amazon \\
  USA \\
  \texttt{douggray@amazon.com} \\
}

\newcommand{\code}[1]{\texttt{\url{#1}}}
\newcommand{\toolcall}[1]{\texttt{#1}}

\begin{document}

\maketitle

\begin{abstract}
Reasoning agents increasingly rely on external tools such as web search to answer complex queries. Reinforcement learning (RL) finetuning algorithms such as GRPO have improved long-form reasoning in text-only language models, particularly for coding and mathematics. Reliable tool use in multimodal agents, however, remains challenging because models must interpret text and images while integrating noisy retrieved evidence, often under sparse outcome-level supervision without explicit verification signals. We present Self-Verification via Reinforcement Learning (SVRL), an RL-only finetuning framework that trains multimodal agents to verify and filter retrieved evidence within their own reasoning traces, reducing reliance on external verifiers at inference time. SVRL also introduces a search-aware penalty that discourages unnecessary tool calls and a query-diversity reward that encourages diverse, well-formed search queries, providing fine-grained feedback on when and what to search. Finetuning Qwen-2.5-VL-7B with SVRL on only 5{,}000 visual question answering examples yields consistent gains in multi-hop VQA generalization and tool efficiency across benchmarks. Overall, SVRL narrows the gap between compact agents and much larger proprietary models while requiring substantially lower training and inference cost.
\end{abstract}

\input{sections/1_intro}

\input{sections/2_related_works}

\input{sections/3_methods}

\input{sections/4_experiments}

\input{sections/5_results}

\input{sections/6_discussion}


\appendix
\input{sections/supplementary}

\bibliographystyle{plainnat}
\bibliography{main}

\end{document}

%% file: sections/1_intro.tex
\section{Introduction} \label{sec:intro}

Multimodal large language models (MLLMs) are evolving from static perception systems into interactive agents that combine vision and text modalities with external tool use. Commercial models such as ChatGPT and Gemini can interpret images and invoke actions such as web search or code execution \cite{gpt4_report,gemini_report}, but their training data, alignment objectives, and tool-use policies are largely proprietary, limiting scientific control and reproducibility. In open-source MLLMs, capability is commonly built through visual instruction tuning \cite{llava_vit,instructblip}, data-driven self-supervision for tool invocation \cite{toolformer}, and preference-based optimization \cite{dpo, christiano2017rlhf, instructgpt, ziegler2019humanpref}. More recently, RL finetuning has re-emerged as a practical route to improving long-horizon reasoning in language models, aided in part by domains with automatic verifiers such as code execution and formal mathematics \cite{deepseekmath,deepseek_r1, rlfinetune_chen2021humaneval, rlfinetune_le2022coderl}. Extending these gains to multimodal tool use is substantially more challenging because retrieved tool outputs are unstructured, and thus rarely admit reliable automatic verifiers.

\begin{figure*}[t]
  \centering

  \begin{subfigure}[t]{0.9\linewidth}
    \centering
    \includegraphics[page=1, width=\linewidth]{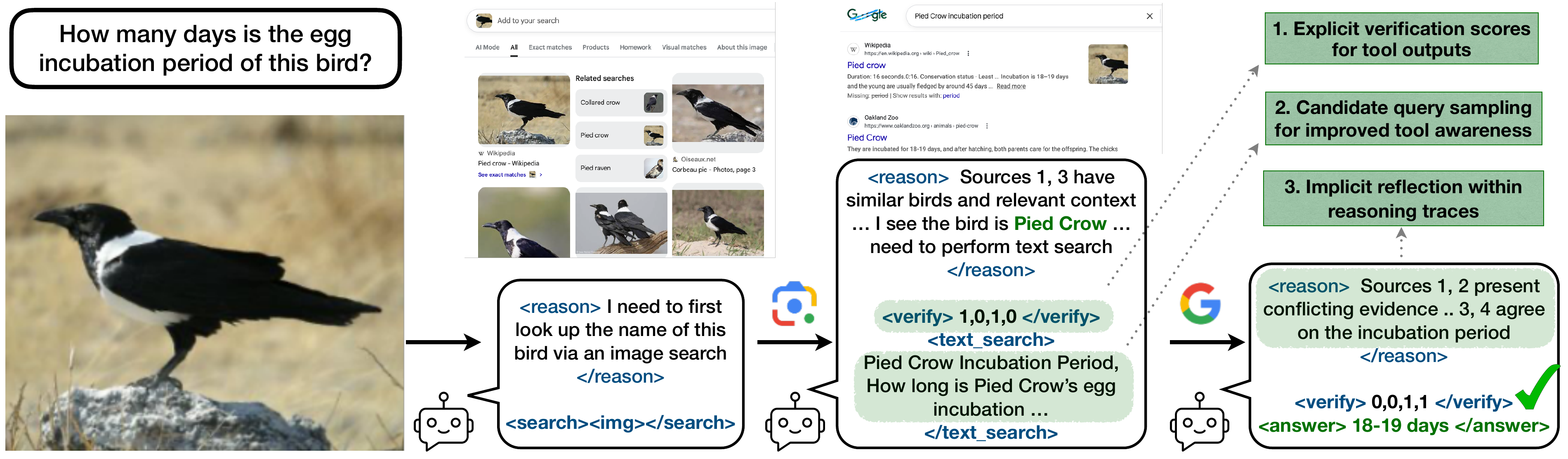}
    \caption{Successful multi-turn reasoning and verified tool use with \textbf{SVRL}.}
    \label{fig:fail_success}
  \end{subfigure}

  \begin{subfigure}[t]{0.33\linewidth}
    \centering
        \includegraphics[page=2, width=\linewidth]{figures/eccv26_figures.pdf}
    \caption{No tool calibration in \textbf{SFT/ Inference} methods.}
    \label{fig:fail_unnecessary}
  \end{subfigure}\hfill
  \begin{subfigure}[t]{0.60\linewidth}
    \centering
    \includegraphics[page=3, width=\linewidth]{figures/eccv26_figures.pdf}
    \caption{Unverified results derail the model towards incorrect reasoning, tool use and final answers in \textbf{Vanilla GRPO}.}
    \label{fig:fail_noisy}
  \end{subfigure}

\caption{Multi-hop tool-augmented VQA example and common pitfalls. The question requires identifying the bird and retrieving its incubation period (18--19 days).}
  \label{fig:failures}
\end{figure*}

To study this setting, we use multi-hop visual question answering (VQA) as a concrete instantiation of tool-augmented multimodal reasoning, where a model answers a natural-language question grounded in an image. In knowledge intensive benchmarks such as InfoSeek and OK-VQA, the image alone is often insufficient to answer a question: the model must recognize visual entities and retrieve missing facts from external sources \cite{infoseek,okvqa,aokvqa,livevqa,mmsearch}. This produces multi-hop reasoning over perception, tool calls, and noisy evidence integration, similar to visual information seeking and active inference processes \cite{avis, chen2023webvln, friston2017activeinference, rao_2023_apc}. Figure~\ref{fig:failures} shows a running example in this regime and highlights representative pitfalls we repeatedly encounter in practice. We discuss them below.

First, agents are often poorly calibrated about \emph{when} search is needed. They may fail to invoke tools when external evidence is necessary or invoke them even when the answer is available from internal knowledge. We analyzed multi-hop reasoning traces produced by MMSearch-R1, an existing GRPO-based finetuning method, on a 2{,}000-example subset of FVQA \cite{mmsearch_r1}. In our analysis, search was required but not invoked in over 10\% of cases, while over 30\% of cases triggered unnecessary search. Figure~\ref{fig:fail_unnecessary} illustrates one such failure mode where a tool call was necessary, but never made. Prior work emphasizes on-demand tool use and the accuracy-cost tradeoff, suggesting that deciding when to search is itself a core part of the problem \cite{search-r1,mmsearch,mmsearch_r1}.

Second, even when search is warranted, tool outputs are noisy and require filtering. Retrieved pages can be irrelevant, outdated, or internally inconsistent. Without an explicit verification mechanism, models may copy the first plausible snippet, ignore disconfirming evidence, or fail to reconcile conflicts (Figure~\ref{fig:fail_noisy}). In our analysis of MMSearch-R1, nearly 45\% of retrieved results were irrelevant. Even when the agent issued an appropriate query and the returned results contained relevant evidence, it answered incorrectly in over 28\% of cases. Prior visual web agents report similar failure patterns under clutter and distractors, often using additional re-ranking or verification stages to mitigate them. These stages add computation, latency, and cost, and can become a deployment bottleneck \cite{webwatcher,deepmmsearch_r1,mmsearch}.

Third, learning reliable tool use is difficult under sparse supervision. Many pipelines provide only an outcome-level reward for the final answer, leaving intermediate decisions (what to search, which result to open, what evidence to trust, when to stop) unlabeled. As a result, distinct failure cases can receive the same terminal penalty, making credit assignment brittle when step annotations are scarce. This sparsity also interacts poorly with GRPO: if all rollouts in a group receive the same reward, the within-group variance collapses and the standardized advantages provide little to no learning signal~\cite{rlfinetune_deng2025grpoCollapse}. Consistent with this, 19.5\% of MMSearch-R1's text search queries are irrelevant, derailing trajectories early and reducing the information in the final GRPO correctness reward.

Motivated by these issues, we introduce \textbf{Self-Verification via Reinforcement Learning} (SVRL), an RL-only finetuning algorithm for tool-augmented multimodal agents. Section~\ref{sec:related} situates SVRL relative to prior work, and Section~\ref{sec:methods} presents the method and training objective. Section~\ref{sec:exp} describes the experimental setup and evaluation benchmarks. In Section~\ref{sec:result}, we evaluate SVRL on both in-distribution and out-of-distribution settings, and examine its behavior through tool-use analyses and qualitative examples. We also study whether SVRL benefits from a larger inference-time search budget by increasing the number of candidate queries available during search. In summary, our main contributions are:
\begin{enumerate}
  \item \textbf{Fine-grained search control.} We augment the GRPO objective with reward terms that discourage unnecessary search and reward informative, diverse query proposals. This provides fine-grained trajectory-level feedback that targets both \emph{when} to search and \emph{what} to search for.
\item \textbf{Self-verification for evidence filtering.} We elicit structured verification scores over retrieved items within the agent's reasoning trace and optimize them as part of the RL objective. These scores make evidence selection and filtering explicit and learnable, and are applied at inference without any external verifier.  
\item \textbf{Stable optimization and test-time query scaling.} We adopt Dr.\ GRPO to stabilize optimization under trajectory-dependent rewards. We further show that increasing the number of candidate search queries at inference improves accuracy, indicating that SVRL supports effective test-time scaling of search.
\end{enumerate}

%% file: sections/2_related_works.tex
\section{Related Work}
\label{sec:related}

\noindent\textbf{Visual Question Answering.}
VQA spans early vision--language architectures and large-scale dataset-driven learning, and has evolved toward settings that require compositional reasoning, reading text, and incorporating external knowledge \cite{vqa,vqav2,clevr,gqa_dataset,textvqa}. Knowledge-intensive and multi-hop benchmarks such as OK-VQA, A-OKVQA, InfoSeek, WebQA, and LiveVQA further stress evidence acquisition and aggregation beyond the image \cite{okvqa,aokvqa,infoseek,webqa,livevqa}. Our work targets this regime: we train on FVQA (as released in MMSearch-R1) and evaluate both in- and out-of-distribution, emphasizing tool-mediated multi-hop behavior rather than single-pass perception \cite{mmsearch_r1,mmsearch}.

\noindent\textbf{LLM finetuning for tool use.}
Modern MLLM pipelines typically start from pretrained transformer language-model backbones \cite{vaswani2023attentionneed, gpt3_brown2020} and acquire interactive behavior through instruction tuning \cite{flan_wei2022,llava_vit,instructblip} and prompting that elicits explicit intermediate reasoning \cite{cot}. For knowledge-intensive QA, retrieval-augmented generation (RAG) injects retrieved context into the model input \cite{rag}; we report a RAG workflow baseline in Table~\ref{tab:main_results} to separate the benefits of always-on retrieval from selective, agentic search. Tool use itself can be induced from data \cite{toolformer,react} and further aligned with preference-based objectives \cite{dpo,instructgpt}. Reinforcement learning finetuning has recently shown strong gains on long-horizon reasoning in language models \cite{ppo,deepseekmath,deepseek_r1}, and concurrent multimodal systems apply RL to improve web search behavior \cite{search-r1,mmsearch_r1,deepmmsearch_r1}. SVRL builds on this direction, but targets calibrated search decisions and explicit evidence filtering under noisy retrieval via fine-grained trajectory rewards that make tool-use errors more distinguishable during optimization.

\noindent\textbf{Verification, reward design, and test-time scaling.}
Reliability is also improved by spending more compute at inference time (e.g., self-consistency and search branching) \cite{self_consistency,tot}, or by adding external re-ranking and verification stages in web-based agents \cite{avis,webwatcher,mmsearch}. Complementary lines study process-level supervision and verifier training for long-horizon reasoning \cite{lets_verify,training_verifiers}, as well as stabilization variants for RL finetuning \cite{dapo,dr_grpo}. Closest to our emphasis on verification signals are recent works that introduce explicit grounding or process-level supervision, but these approaches differ in source of supervision, target task and modality, or where verification is applied in the pipeline \cite{vigorl_sarch2025,vstar_du2023,freeprocreward_chen2024}. SVRL focuses on making verification and filtering behaviors learnable from compact multimodal rollouts while keeping inference free of external verifiers, and we empirically study how this interacts with GRPO-style learning dynamics under trajectory-level rewards.

%% file: sections/3_methods.tex
\section{Self-Verification via Reinforcement Learning} \label{sec:methods}

In this section, we present SVRL, a Reinforcement Learning based finetuning framework for multimodal agents. We formalize the setting and objective, introduce two algorithmic modifications for trajectory-level rewards and self-verification, and discuss practical considerations for stable training.

\subsection{Preliminaries}
\label{sec:prelims}

\paragraph{\textbf{Visual Question Answering (VQA)}.}
We consider a VQA dataset $\mathcal{D}$ of triples $(I,q,y^*)$, where $I$ is an image, $q$ is a natural-language question about the image, and $y^*$ is the ground-truth answer. While some instances can be answered directly from the image, we focus on multi-hop questions that require external knowledge and tool use (see Figure~\ref{fig:failures}) \cite{vqa,okvqa,aokvqa,infoseek}.

\paragraph{\textbf{Multimodal reasoning as policy optimization}.}
Our primary goal is to improve a multimodal agent’s tool-use and reasoning capabilities at inference time. Given an input $(I,q)$, we view the model as a stochastic policy $\pi_\theta$ that generates a trajectory $\tau$ consisting of $T$ intermediate reasoning steps $s_{1:T}$ and a final answer $y$, i.e., $\tau=(s_{1:T},y)$. The trajectory likelihood factorizes as
\begin{equation}
\pi_\theta(\tau \mid I,q)=\Big(\prod_{t=1}^{T}\pi_\theta(s_t\mid I,q,s_{<t})\Big)\,\pi_\theta(y\mid I,q,s_{\le T}).
\label{eq:traj}
\end{equation}
Each step $s_t=(l_t,a_t,p_t)$ follows a structured format and may contain (i) a natural-language reasoning $l_t$ within \toolcall{<reason>...</reason>} tags and (ii) an action $a_t$ (e.g., a tool invocation or emitting the final answer) with optional tool parameters $p_t$. A representative multi-step reasoning process can be seen in Figure~\ref{fig:fail_success}. We follow the tool set and formatting conventions of MMSearch-R1 throughout this paper for simplicity \cite{mmsearch_r1}.

\paragraph{\textbf{Group Relative Policy Optimization (GRPO)}.}
We finetune $\pi_\theta$ with RL to maximize a trajectory-level reward $R(\tau)$, regularized by a KL penalty to a fixed reference policy $\pi_{\theta_{\mathrm{ref}}}$:
\[
\max_{\theta}\; \mathbb{E}_{(I,q,y^*)\sim\mathcal{D}}
\Big[
\mathbb{E}_{\tau\sim \pi_\theta(\cdot\mid I,q)}[R(\tau)]
-\beta\,D_{\mathrm{KL}}\!\big(\pi_\theta(\cdot\mid I,q)\,\|\,\pi_{\theta_{\mathrm{ref}}}(\cdot\mid I,q)\big)
\Big].
\]
GRPO is a PPO-style optimizer that estimates advantages from groups of on-policy rollouts without a learned value function \cite{ppo,deepseekmath}. For each $(I,q)$, we sample a group $\tau^{1:G}$ using a behavior policy $\pi_{\theta_{\mathrm{old}}}$, compute standardized group advantages $\hat{A}_i$ from $\{R(\tau^{(i)})\}_{i=1}^G$, and apply the clipped surrogate $\mathrm{PPOClip}(r,A)=\min(rA,\operatorname{clip}(r,1-\epsilon,1+\epsilon)A)$ with an importance ratio $r_i(\theta)=\pi_\theta(\tau^{(i)}\!)/\pi_{\theta_{\mathrm{old}}}(\tau^{(i)}\!)$:
\begin{equation}
\max_{\theta}\; \mathbb{E}_{\mathcal{D}}\;
\mathbb{E}_{\tau^{1:G}}
\left[
\frac{1}{G}\sum_{i=1}^{G}\mathrm{PPOClip}\!\big(r_i(\theta),\hat{A}_i\big)
-\beta\,D_{\mathrm{KL}}\!\big(\pi_\theta\,\|\,\pi_{\theta_{\mathrm{ref}}}\big)
\right]
\label{eq:grpo}
\end{equation}

\subsection{When and What to Search?}
\label{sec:when}

Humans typically decide whether to search and how to phrase a query by trading off expected information gain against time and effort, consistent with ideas in rational metareasoning and information foraging \cite{rlfinetune_russell1991,rlfinetune_pirolli1999}. We reflect this intuition by augmenting the GRPO maximization objective with search-aware and query-diversity rewards, so trajectory returns depend explicitly on \emph{when} search is invoked and on the quality of the proposed queries. We begin from the RL finetuning setup in MMSearch-R1 \cite{mmsearch_r1}, which trains a multimodal agent to use text and image search with a reward of the form $ r_{\mathrm{base}}=(1-\alpha)\,r_{\mathrm{acc}}\cdot r_{\mathrm{search}} \;+\; \alpha\,r_{\mathrm{fmt}}.$ Here, $r_{\mathrm{acc}}$ is exact-match accuracy against the ground-truth answer, $r_{\mathrm{fmt}}$ enforces a valid action format (common in RL finetuning of reasoning models) \cite{deepseek_r1}, and $r_{\mathrm{search}}$ discourages excessive tool calls to reduce cost and latency.

\paragraph{\textbf{Tool calibration (when to search)}.}
A uniform step-style penalty is well motivated when learning a policy from scratch (RL navigation, manipulation, etc), but it can be sub-optimal when finetuning a pretrained model that already answers a subset of questions without tools. Penalizing search uniformly can also suppress necessary tool use, since it does not distinguish avoidable search from required search. We therefore construct a \emph{search-aware} factor using the pretrained model as a weak competence prior. Concretely, for each training input $(I,q)$ we run a single forward pass on the model without tools to obtain $\hat{y}$; if $\hat{y}=y^*$, we label the instance as \toolcall{search\_free}. During RL rollouts, we downweight trajectories that invoke a search tool on a \toolcall{search\_free} instance:
\[
r_{\mathrm{aware}}=
\begin{cases}
\lambda_{\mathrm{sa}} & \text{if \toolcall{search\_free} and a search tool is called},\\
1 & \text{otherwise}.
\end{cases}
\]
In our experiments, we set $\lambda_{\mathrm{sa}}=0.1$ and replace the baseline search penalty with $r_{\mathrm{aware}}$. This targets avoidable tool calls while preserving the incentive to search when needed. Empirically, it improves both the accuracy--search tradeoff and overall tool calibration (Table~\ref{tab:grpo_ablations}, Figure~\ref{fig:analysis}).
\paragraph{\textbf{Query diversity (what to search)}.}
We additionally find that text-search queries produced by the agent are often underspecified, even when search is necessary. To encourage better query formulation, we prompt the model to propose $k$ candidate queries and randomly execute one of the valid, unique queries. Let $\hat{k}\le k$ denote the number of unique queries that are produced in the required format. We define a query-count factor
\[
r_{\text{count}}=
\begin{cases}
1, & \text{if the rollout makes no text-search call},\\
\max(\epsilon,\hat{k}/k), & \text{otherwise},
\end{cases}
\]
where $\epsilon>0$ is a small floor to avoid degenerate zero reward when no valid queries are produced. We incorporate $r_{\text{count}}$ multiplicatively into the task term of the reward, so rollouts are explicitly differentiated by the number of usable query proposals. This encourages the agent to generate multiple well formed, diverse queries, improving the specificity of search intent and downstream evidence retrieval.
\subsection{Verifying Search Results} \label{sec:self}

\begin{figure}[t]
  \centering
  \includegraphics[page=4, width=0.9\linewidth]{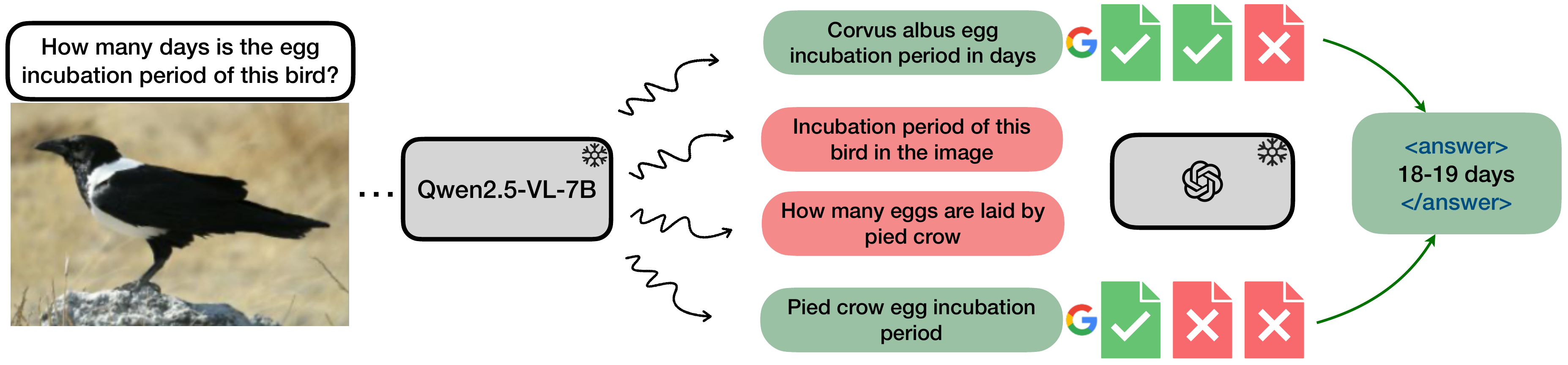}
  \caption{\textbf{Test Time Verification}. We demonstrate the need for verification of search queries and tool outputs via an external verifier. GPT-5 filters incorrect search queries and tags noisy web page snippets. With this verification process and no additional model finetuning, we observe improvements in multi-hop VQA performance.}
  \label{fig:ext_verify}
\end{figure}

Web retrieval is noisy: search results can be irrelevant, outdated, or internally inconsistent. Drawing on signal-detection and source-monitoring views of validation under uncertainty \cite{rlfinetune_green1966,rlfinetune_johnson1993}, we make evidence filtering explicit by training the agent to score the usefulness of retrieved items and suppress misleading snippets. This targets a common failure of GRPO-finetuned compact agents, which often over-trust early snippets and aggregate brittle evidence, consistent with prior web-based multimodal agents \cite{mmsearch,webwatcher,deepmmsearch_r1}.

\paragraph{\textbf{Test Time Verification}.} To quantify the benefit of denoising, we introduce an external verifier at inference time. Specifically, we use a stronger text-only model (GPT-5) to filter both query candidates and retrieved snippets without access to the ground-truth answers. For each input $(I,q)$, we generate multiple candidate text queries from the base agent (via repeated decoding with temperature $=1$), and ask the verifier to retain only the queries that are likely to retrieve relevant evidence. We then run text search using the retained queries and ask the verifier again to select useful snippets given the original $(I,q)$ and the retrieved context (Figure~\ref{fig:ext_verify}). This test-time filtering consistently improves final accuracy while keeping the answering model fixed, which isolates the error source to noisy tool outputs and weak evidence selection (See Table~\ref{tab:grpo_ablations}).

\begin{figure}[t]
  \centering
  \includegraphics[page=5, width=\linewidth]{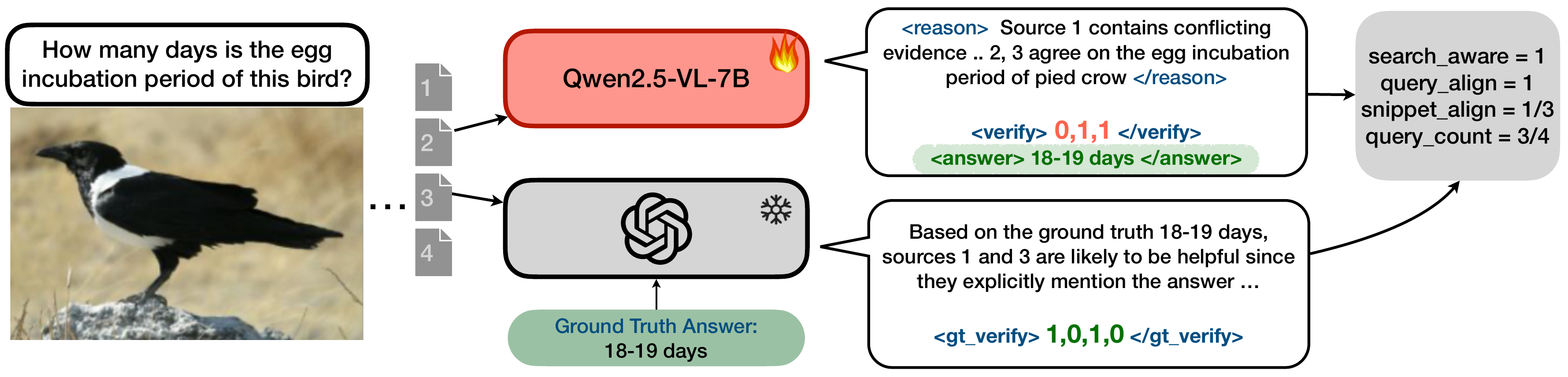}
  \caption{Self-Verification via oracle feedback during finetuning. GPT-5 is used as the oracle verifier that receives the same web snippets as the model and additionally the ground truth answer. The verifier then annotates each web snippet based on the utility towards inferring the correct answer. This verifier is discarded during inference.}
  \label{fig:self_verify}
\end{figure}

\paragraph{\textbf{Self-Verification}.}
Test-time verification improves accuracy but is computationally expensive and reduces the deployment advantage of compact agents. Our goal is to recover these gains without any external verifier at inference time. We do so by eliciting self-verification inside the agent's reasoning trace (Figure~\ref{fig:self_verify}) and optimizing it directly during RL finetuning. Concretely, for each input $(I,q)$ the agent (i) proposes multiple candidate text-search queries and (ii) assigns a structured binary usefulness vector to retrieved items, e.g., \toolcall{<verify> 0,1,0,1 </verify>}, together with a short natural-language justification.

To avoid reward-hacking from additive shaping terms (Section~\ref{sec:robust}), we keep verification supervision modular and apply it only when the final answer is correct. Let $r_{\mathrm{aware}}$ and $r_{\mathrm{count}}$ denote the search-calibration factors from Section~\ref{sec:when}. We introduce two additional alignment factors, both supervised at training time by an external verifier that scores queries and snippets: $r_{\mathrm{qalign}}$ measures agreement between the agent's query proposals and verifier labels of which proposed queries are useful, and $r_{\mathrm{salign}}$ measures agreement between the agent's snippet-level self-scores and verifier labels of which retrieved items contain answer-supporting evidence. We combine these terms into a single SVRL factor,
\begin{equation}
r_{\mathrm{svrl}}
=
r_{\mathrm{aware}}\cdot r_{\mathrm{count}}\cdot r_{\mathrm{qalign}}\cdot r_{\mathrm{salign}}.
\label{eq:r_svrl}
\end{equation}
Let $r_{\mathrm{acc}}\in\{0,1\}$ denote exact-match correctness and $r_{\mathrm{fmt}}\in[0,1]$ denote the format reward. Our final trajectory reward is
\begin{equation}
r(\tau)
=
(1-\alpha)\,r_{\mathrm{acc}}(\tau)\,r_{\mathrm{svrl}}(\tau)
+\alpha\,r_{\mathrm{fmt}}(\tau).
\label{eq:svrl_reward}
\end{equation}

During training, we obtain the query and snippet labels using a verifier model with access to the ground-truth answer $y^*$, which is necessary to robustly handle aliases and paraphrases beyond string matching. We then finetune the agent so that its self-verification scores align with these labels. Figure~\ref{fig:self_verify} illustrates the SVRL scores for a sample reasoning trace. The verifier is discarded at inference time, where the agent must perform verification and filtering on its own. Finally, we incorporate Equation~\ref{eq:svrl_reward} into the KL-regularized GRPO objective by replacing the trajectory reward $R(\tau)$ with our $r(\tau)$:
\begin{equation}
\max_{\theta}\; \mathbb{E}_{(I,q,y^*)\sim\mathcal{D}}
\Big[
\mathbb{E}_{\tau\sim \pi_\theta(\cdot\mid I,q)}[r(\tau)]
-\beta\,D_{\mathrm{KL}}\!\big(\pi_\theta(\cdot\mid I,q)\,\|\,\pi_{\theta_{\mathrm{ref}}}(\cdot\mid I,q)\big)
\Big].
\label{eq:svrl_grpo_obj}
\end{equation}
This yields a structured, fine-grained, trajectory-level learning signal that targets query quality and evidence filtering, improving tool integration and accuracy in Table~\ref{tab:grpo_ablations} and Table~\ref{tab:main_results}. Algorithm~\ref{alg:svrl} summarizes the SVRL finetuning algorithm discussed in this section.

\begin{algorithm}[t]
\caption{SVRL finetuning}
\label{alg:svrl}
\begin{algorithmic}[1]
\Require Dataset $\mathcal{D}$, system prompt $\mathcal{S}$, train-time verifier $\mathcal{G}$, policy $\pi_\theta$, search tools $\mathcal{T}$, format weight $\alpha$, group size $G$

\State \textbf{Self-labeling}.
\ForAll{$(I,q,y^*) \in \mathcal{D}$}
  \State $\hat{y} \gets \pi_\theta(\mathcal{S}, I, q)$ \Comment{single pass; no tool calls}
  \State $\ell(I,q) \gets \mathbb{I}[\hat{y}=y^*]$ \Comment{$1$ iff \code{search_free}}
\EndFor

\State \textbf{RL finetuning with Dr.GRPO}.
\While{not converged}
  \State Sample minibatch $\mathcal{B}\subset\mathcal{D}$
  \ForAll{$(I,q,y^*) \in \mathcal{B}$}
    \State Sample rollouts $\{\tau^{(i)}\}_{i=1}^{G} \sim \pi_{\theta_{\mathrm{old}}}(\cdot \mid \mathcal{S},I,q;\mathcal{T})$
    \For{$i=1$ to $G$}
      \State Parse $\tau^{(i)} \rightarrow (y, p, v)$ \Comment{answer, tool params, self-verification}
      \State $v^{*} \gets \mathcal{G}(I,q,y^*,p)$ \Comment{train-time verification labels}

\State $r_{\texttt{svrl}} \gets \code{search_aware}(\ell(I,q),\tau^{(i)}) \cdot \code{query_count}(\tau^{(i)})$ \Comment{Section~\ref{sec:when}}
\State $r_{\texttt{svrl}} \gets r_{\texttt{svrl}} \cdot \code{query_align}(v,v^*) \cdot \code{snippet_align}(v,v^*)$ \Comment{Section~\ref{sec:self}}
     
     \State $r^{(i)} \gets (1-\alpha) \cdot \code{acc_score}\cdot r_{\texttt{svrl}} + \alpha \cdot \code{format_score}$
    \EndFor
    \State Update $\theta$ using $\{(\tau^{(i)}, r^{(i)})\}_{i=1}^{G}$ \Comment{group advantages + Dr.GRPO objective}
  \EndFor
\EndWhile
\end{algorithmic}
\end{algorithm}

\subsection{Practical Considerations}
\label{sec:robust}

\paragraph{\textbf{Reward diversity for GRPO.}}
GRPO estimates group-relative advantages by normalizing rewards within a rollout group for each input $(I,q)$ (subtracting the group mean and dividing by the group standard deviation) \cite{deepseekmath}. If all rollouts in a group receive the same reward, there is no relative preference signal and the resulting update becomes uninformative; if the within-group reward dispersion is very small, the normalization is poorly conditioned and gradients can become noisy. With the baseline reward $r_{\mathrm{base}}$ (Section~\ref{sec:when}), we often observe near-tied groups because trajectories share the same correctness and similar tool-use patterns. In contrast, $r_{\mathrm{svrl}}$ varies with query formulation, retrieved snippets, and verification outcomes, increasing within-group reward dispersion and yielding more informative advantages. Empirically, this lets us use a smaller group size ($G=4$) than prior work (MMSearch-R1 uses $G=8$), improving throughput on a single 8$\times$A100 node.

\begin{figure}[t]
  \centering
  \includegraphics[width=\linewidth]{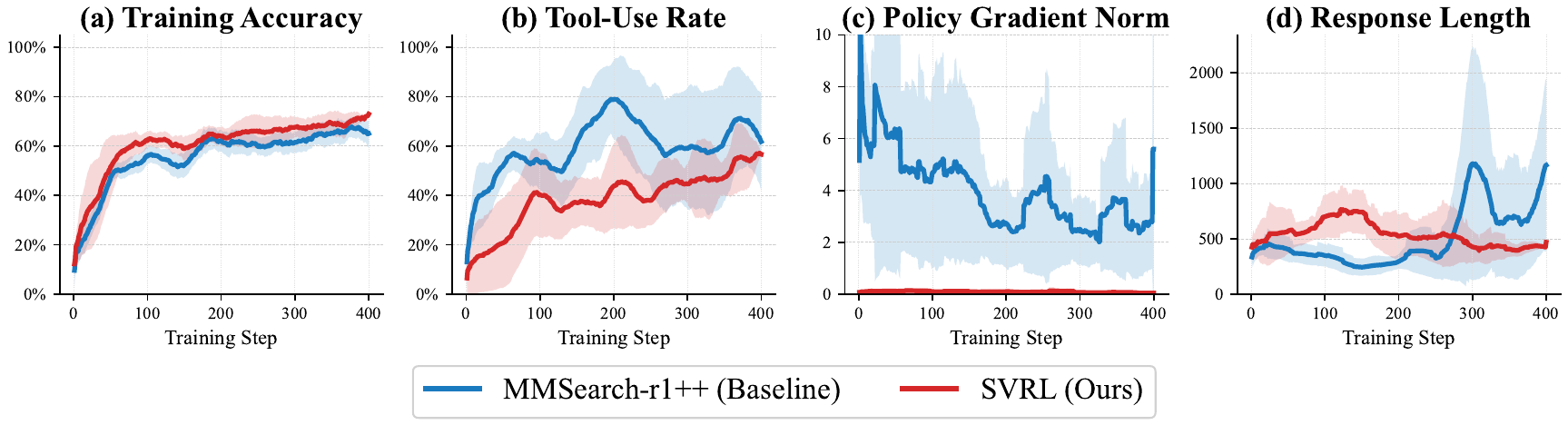}
  \caption{\textbf{SVRL yields more stable RL finetuning dynamics.}
Mean over 3 seeds with $\pm 1$ standard deviation shown. All runs use 400 steps, batch size 32, group size 4, and multi-turn tool-use trajectories. SVRL is compared against MMSearch-R1++ (MMSearch-R1 reproduced with ReAct-style prompting \cite{mmsearch_r1,react}). (a) training performance, (b) search ratio, (c) policy gradient norm, and (d) response length.}
  \label{fig:drgrpo}
\end{figure}

\paragraph{\textbf{Dr.\ GRPO for training stability.}}
In early experiments, vanilla GRPO exhibited unstable optimization, including intermittent spikes in actor gradient norms and drift in response length (Figure~\ref{fig:drgrpo}). We therefore adopt Dr.\ GRPO, which removes normalization factors in the update (for example, standard-deviation and length normalization) that can amplify gradients under trajectory-dependent rewards \cite{dr_grpo}. We use Dr.\ GRPO for all SVRL variants reported in Table~\ref{tab:grpo_ablations} and Table~\ref{tab:main_results}.

\paragraph{\textbf{LLM-judge rewards and reward hacking.}}
We also tried replacing exact-match correctness with an LLM-judge reward during training (Qwen3-32B comparing $y$ to $y^*$). Although judge rewards could tolerate aliases, they were vulnerable to reward hacking: the policy increased answer verbosity and began to output hedged or multi-answer responses that judges sometimes scored favorably despite the answers themselves being non-committal. This is consistent with known judge failure modes \cite{instructgpt,mtbench,geval,llm_judge_survey}. We therefore use exact-match correctness for training, and reserve LLM judges for evaluation-time scoring (following the benchmark protocol). For reward shaping, we found additive mixtures of shaping terms to be easier to exploit (partial credit despite incorrect answers), so we gate fine-grained signals by correctness via the multiplicative factor $r_{\mathrm{acc}} . r_{\mathrm{svrl}}$. Another instance of reward hacking occurred in the query-alignment ablation without the search-aware factor. Although query alignment is gated by correctness, the model learned to invoke search on nearly every example to maximize the query-alignment reward while still producing correct answers. Adding the search-aware factor reduced unnecessary search, while the full self-verification objective further improved both accuracy and tool efficiency (Table~\ref{tab:grpo_ablations}).

%% file: sections/4_experiments.tex
\section{Experimental Setup} \label{sec:exp}

\begin{table}[t]
\caption{Ablations on finetuning with GRPO. 
Acc (\%) denotes LLM-as-judge accuracy. 
SR (\%) denotes search ratio. Details in Section~\ref{sec:result}.}
\label{tab:grpo_ablations}
\centering
\small
\setlength{\tabcolsep}{8pt}
\begin{tabular}{l|cc|cc}
\toprule
\multicolumn{1}{c|}{\multirow{2}{*}{Model}}
& \multicolumn{2}{c|}{FVQA-test}
& \multicolumn{2}{c}{InfoSeek} \\
\cmidrule(lr){2-3}
\cmidrule(lr){4-5}
& Acc & SR & Acc & SR \\
\midrule

MMSearch-R1  & 51.1 & 64.2 & 49.8 & 63.5 \\
MMSearch-R1++ & 56.4 & 80.3 & 55.1 & 59.7 \\
SVRL-query-align & 57.7 & 97.6 & 58.2 & 98.0 \\
SVRL-llm-judge & 58.1 & 36.7 & 58.0 & 50.6 \\
SVRL-search-aware & 59.9 & 80.3 & 57.0 & 85.8 \\
SVRL-ttv & 60.6 & 80.3 & 58.3 & 85.2 \\
SVRL-self-verification & 64.2 & 61.2 & 64.1 & 59.2 \\

\bottomrule
\end{tabular}
\end{table}

\paragraph{\textbf{Models}.}
Our primary experiments use the base Qwen2.5-VL-7B-Instruct backbone which we finetune with our method and evaluate on all benchmarks~\cite{qwen25vl}. We further re-implement and finetune with the baseline GRPO method presented in MMSearch-R1 and an improved method MMSearch-R1++ that replaces the original system prompt with a ReAct based prompt~\cite{mmsearch_r1}. Additionally, for strong comparison to modern LLMs, we also report inference-time results for GPT-4o across the datasets used.

\paragraph{\textbf{Datasets}.} 
We train Qwen-2.5-VL-Instruct with our algorithm on the Factual VQA (FVQA) dataset. FVQA contains a mix of direct-answer questions and multi-hop questions that require tools, which helps RL avoid degenerating into tool use on every example. Following Wu et al., we use 5{,}000 examples for training and reserve 1{,}800 for testing. We evaluate on FVQA and InfoSeek as in-distribution benchmarks, since FVQA is sampled from the broader InfoSeek collection and shares the same underlying data and task structure. To assess robustness, we additionally evaluate out-of-distribution on LiveVQA, MMSearch and SimpleVQA datasets, which differ in query style and retrieval conditions and therefore test whether the learned behavior transfers beyond the training distribution \cite{mmsearch_r1, infoseek, mmsearch, livevqa, simplevqa}.

\paragraph{\textbf{Implementation details}.}
We use an LLM judge (GPT-5) to determine final answer correctness when computing accuracy, following the evaluation protocol and judge prompt presented in \cite{mmsearch_r1}. This choice provides a scalable and consistent correctness signal across benchmarks where answers may have minor surface-form variations (e.g., aliases, physical units, date-time formats), while keeping our reported numbers directly comparable to prior work. We train our model with a ReAct-style system prompt to structure tool use and intermediate reasoning, and we include the full prompt templates in the appendix \cite{react}. For tool access, we use Google Image Search and Google text search via a third party API. To reduce redundant and expensive API calls, we implement a caching layer for text and image search results on DynamoDB. All training and inference experiments are run on 8$\times$A100 GPUs using PyTorch, Ray and veRL framework.

%% file: sections/5_results.tex
\begin{table*}[t]
\caption{\textbf{Performance across benchmarks.} Acc (\%) denotes answer accuracy and SR (\%) denotes search ratio. SVRL improves accuracy across all datasets and achieves a better accuracy-search tradeoff, indicating more effective multi-turn reasoning and tool use. Details in text.}
\label{tab:main_results}
\centering
\small
\setlength{\tabcolsep}{4pt} 
\begin{tabular}{l|cc|cc|cc|cc|cc}
\toprule
\multicolumn{1}{c|}{\multirow{2}{*}{Model}}
& \multicolumn{2}{c|}{FVQA-test}
& \multicolumn{2}{c|}{InfoSeek}
& \multicolumn{2}{c|}{MMSearch}
& \multicolumn{2}{c|}{LiveVQA}
& \multicolumn{2}{c}{SimpleVQA} \\
\cmidrule(lr){2-3}
\cmidrule(lr){4-5}
\cmidrule(lr){6-7}
\cmidrule(lr){8-9}
\cmidrule(lr){10-11}
& Acc & SR
& Acc & SR
& Acc & SR
& Acc & SR
& Acc & SR \\
\midrule

\multicolumn{11}{c}{\textbf{Direct Answer}} \\
\midrule
Qwen-2.5-VL-7B  & 26.7 & 0.0 & 20.1 & 0.0 & 12.8 & 0.0 & 17.8 & 0.0 & 38.4 & 0.0 \\
Qwen-2.5-VL-32B & 24.7 & 0.0 & 25.8 & 0.0 & 15.7 & 0.0 & 18.7 & 0.0 & 40.1 & 0.0  \\
Qwen-2.5-VL-72B & 27.1 & 0.0 & 28.0 & 0.0 & 15.7 & 0.0 & 20.1 & 0.0 & 42.2 & 0.0 \\
GPT-4o          & 41.7 & 0.0 & 42.7 & 0.0 & 22.2 & 0.0 & 26.9 & 0.0 & 46.6 & 0.0 \\

\midrule
\multicolumn{11}{c}{\textbf{RAG Workflow}} \\
\midrule
Qwen-2.5-VL-7B  & 51.6 & 100 & 53.7 & 100 & 52.2 & 100 & 48.0 & 100 & 51.6 & 100  \\
Qwen-2.5-VL-32B & 57.0 & 100 & 56.8 & 100 & 57.9 & 100 & 49.6 & 100 & 54.5 & 100 \\
Qwen-2.5-VL-72B & 62.2 & 100 & 59.4 & 100 & 59.6 & 100 & 56.0 & 100 & 61.0 & 100 \\
GPT-4o          & 66.0 & 100 & 59.1 & 100 & 62.5 & 100 & 59.6 & 100 & 63.4 & 100 \\

\midrule
\multicolumn{11}{c}{\textbf{Adaptive Search}} \\
\midrule
MMSearch-R1++     & 56.4 & 80.3 & 55.1 & 59.7 & 53.8 & 88.5 & 48.4 & 76.2 & 57.4 & 42.5 \\
\textbf{SVRL-full-7B} & \textbf{65.3} & \textbf{61.4} & \textbf{64.7} & \textbf{52.3} & \textbf{60.3} & \textbf{82.0} & \textbf{57.8} & \textbf{48.6} & \textbf{58.3} & \textbf{27.0} \\

\bottomrule
\end{tabular}
\end{table*}

\section{Results} \label{sec:result}

\paragraph{\textbf{Ablations}.}
Table~\ref{tab:grpo_ablations} isolates the impact of each component. \textbf{MMSearch-R1} reproduces the original recipe~\cite{mmsearch_r1}. \textbf{MMSearch-R1++} strengthens this baseline by switching to a ReAct-style prompt and removing the KL penalty and entropy coefficient~\cite{mmsearchr1_blog}. \textbf{SVRL-llm-judge} adds DrGRPO but replaces exact-match with an LLM judge reward. Here, accuracy improves slightly, but answers become longer and less decisive, consistent with reward hacking. \textbf{SVRL-query-align} adds query alignment without the search-aware factor. It improves accuracy, but the model invokes search on nearly every example to maximize query-alignment reward. \textbf{SVRL-search-aware} adds the calibration term from Section~\ref{sec:when}, improving accuracy by discouraging avoidable search while preserving search when needed. \textbf{SVRL-ttv} applies external test-time verification to filter query candidates and snippets, yielding a large gain that isolates evidence selection as a dominant failure mode. \textbf{SVRL-self-verification} replaces the external verifier with our train-time self-verification objective (Section~\ref{sec:self}), recovering most of the denoising benefit without verifier dependence at inference.

\begin{figure}[t]
  \centering
  \includegraphics[width=0.8\linewidth]{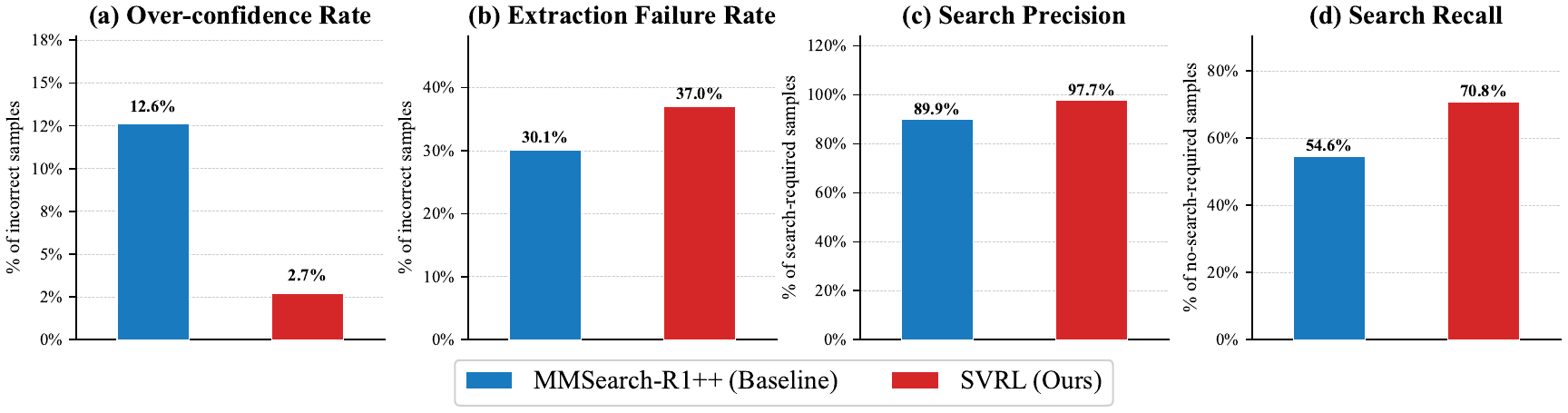}
  \caption{\textbf{Tool-use analysis.} Aggregated FVQA and InfoSeek statistics.}
  \label{fig:analysis}
\end{figure}

\paragraph{\textbf{VQA generalization}.}
Table~\ref{tab:main_results} compares three regimes: \emph{Direct Answer} (single-pass prediction from $(I,q)$), \emph{RAG Workflow} (retrieval for every instance; SR$=100$), and \emph{Adaptive Search} (RL-finetuned agents that decide when to search). SVRL-full-7B consistently improves over the strongest adaptive-search baseline, MMSearch-R1++, by roughly 8 points on FVQA-test and 9 points on InfoSeek while also reducing search ratio, indicating a better accuracy-cost tradeoff. SVRL-full-7B approaches GPT-4o on FVQA-test (65.3 vs.\ 66.0) despite operating in the 7B regime. On SimpleVQA, gains are smaller, which suggests that search behavior may depend on tool parameters such as query language. We also note that time-sensitive or under specified questions can introduce irreducible evaluation noise. Additional training and inference results, and tool use analysis are provided in the Supplementary Materials.

\begin{figure}[t]
  \centering
  \includegraphics[width=\linewidth]{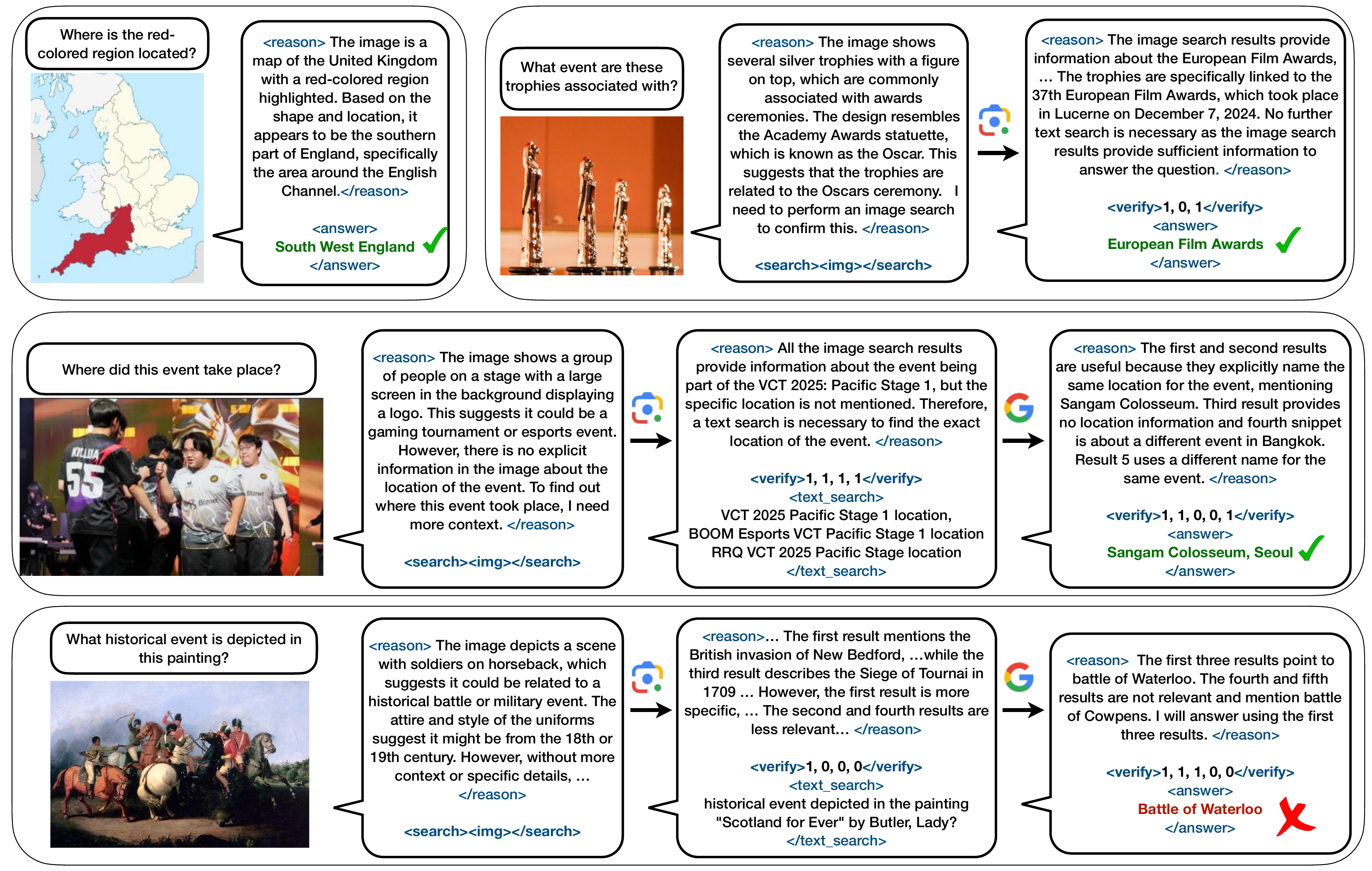}
  \caption{\textbf{Qualitative examples}. FVQA-test cases for trained SVRL model. Top-left shows an example that did not require search. Top-right shows a sample that required only image search. Bottom two rows show multi-turn reasoning over multiple tools - one with a correct and the other with an incorrect answer.}
  \label{fig:qualitative}
\end{figure}

\paragraph{\textbf{Analysis}.}
Figure~\ref{fig:qualitative} shows qualitative multi-hop examples from the SVRL-finetuned agent. We observe that SVRL induces more robust reasoning traces that compare and reconcile evidence across retrieved results, even though we do not supervise free-form critique text directly. This behavior is consistent with the verification-alignment rewards, which make evidence filtering explicit during optimization. Figure~\ref{fig:analysis} summarizes the corresponding changes in tool behavior: SVRL reduces over-confidence failures where search is omitted despite being required (Figure~\ref{fig:analysis}(a)) and improves search precision and recall under the \toolcall{search\_free} annotations (Figures~\ref{fig:analysis}(c) and (d)). Finally, Figure~\ref{fig:analysis}(b) reports the conditional rate at which relevant evidence appears in retrieved content given an incorrect answer, isolating evidence utilization errors from retrieval quality.

\paragraph{\textbf{Test-time scaling}.}
\begin{wrapfigure}{r}{0.32\linewidth}
  \centering
  \includegraphics[width=\linewidth]{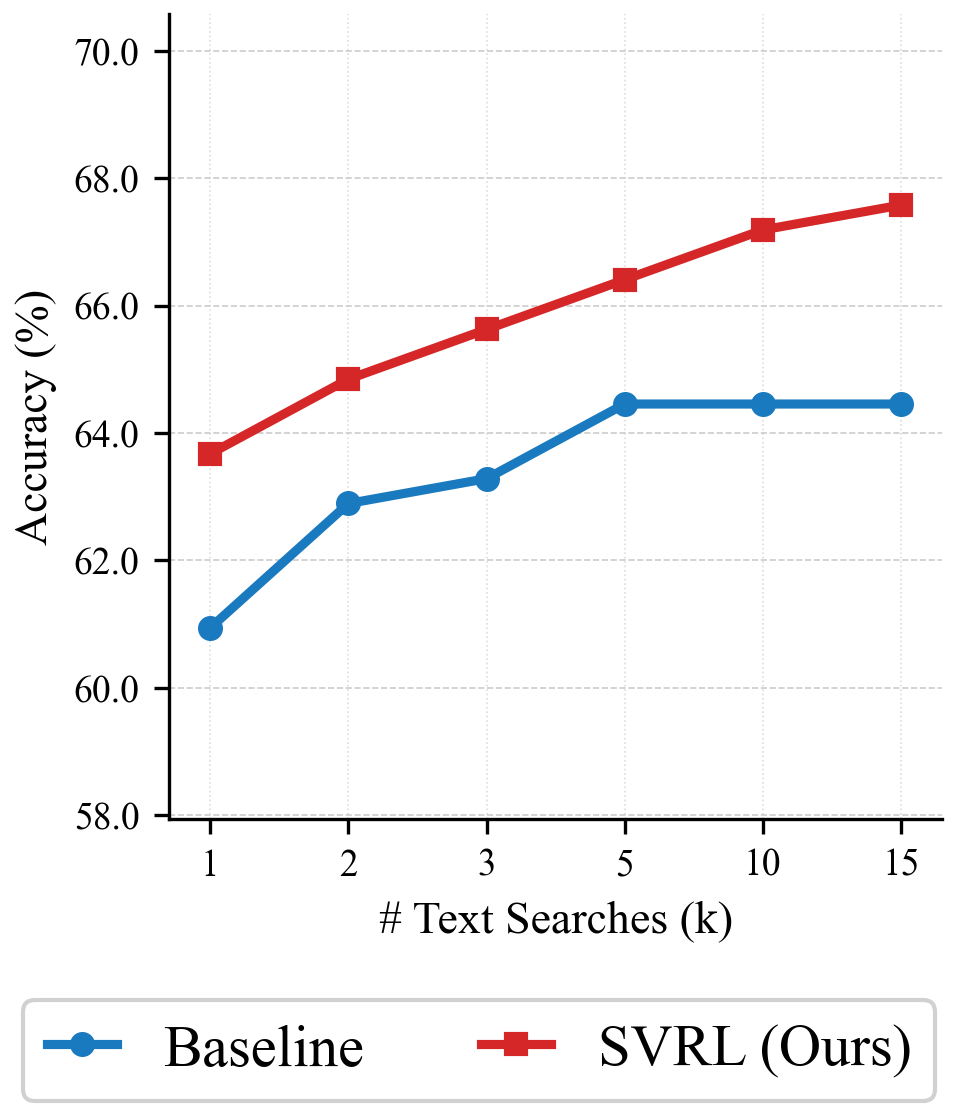}
  \caption{\textbf{Test-time scaling}. SVRL benefits from additional search budget.}
  \label{fig:tts_accuracy}
\end{wrapfigure}
A practical benefit of self-verification is improved \emph{test-time scaling}: allocating more inference-time computation (e.g., sampling, search, and verification) can increase accuracy under a larger compute budget, as studied in self-consistency voting, tree-based search over reasoning paths, and self-evaluation guided beam search~\cite{self_consistency,tot,segb_xie2023}. We run preliminary scaling experiments by increasing the text search budget at inference. Concretely, we evaluate on a 256-example subset of FVQA-test and allow up to 15 parallel text searches per question, producing multiple candidate answers that we aggregate by simple majority vote\footnote{See Supplementary for details}. Figure~\ref{fig:tts_accuracy} reports accuracy as a function of the allowed number of parallel searches. SVRL continues to benefit from additional search budget, whereas the baseline (MMSearch-R1++) saturates after roughly 5 searches, consistent with weaker query diversity and less reliable evidence filtering. Because SVRL proposes multiple queries within a single reasoning step, it also reduces the need for repeated full trajectory sampling, improving wall-clock efficiency. We leave stronger aggregation rules and more structured search procedures for future work.

%% file: sections/6_discussion.tex
\section{Conclusion} \label{sec:conclusion}

We presented SVRL, an RL-only finetuning framework that improves tool augmented multi-hop VQA in compact multimodal agents by eliciting self-verification within their reasoning traces. SVRL turns evidence selection into a learnable trajectory signal by coupling calibrated search decisions and query diversity with verifier-aligned supervision for filtering retrieved content during training, while requiring no external verifier at inference. Across FVQA-test and InfoSeek, SVRL consistently improves accuracy and the accuracy--search tradeoff over prior GRPO-based baselines, and narrows the gap to substantially larger proprietary models. Ablations indicate that search calibration and self-verification drive most of the gains, and test-time verification quantifies the benefit of denoising noisy retrieval. Limitations include dependence on web tools and their failure modes, reliance on a training-time judge whose labels may introduce bias, and sensitivity to prompt and tool formatting. Future work will broaden SVRL to richer tool suites and modalities, reduce judge dependence via stronger self-supervised verification signals, and improve robustness under retrieval distribution shift and evolving web content.

%% file: sections/supplementary.tex

\clearpage
\begingroup

\edef\suppSavedSection{\number\value{section}}
\edef\suppSavedSubsection{\number\value{subsection}}
\edef\suppSavedSubsubsection{\number\value{subsubsection}}
\edef\suppSavedFigure{\number\value{figure}}
\edef\suppSavedTable{\number\value{table}}
\edef\suppSavedEquation{\number\value{equation}}


\definecolor{suppblue}{RGB}{34,102,153}
\definecolor{suppred}{RGB}{180,55,55}

\newcommand{\suppsectionentry}[3]{%
  \noindent
  \makebox[2.8em][l]{%
    \hyperref[#3]{\textcolor{suppred}{\bfseries #1}}%
  }%
  \hyperref[#3]{\textcolor{suppred}{\bfseries #2}}%
  \nobreak\leaders\hbox to 0.45em{\hss.\hss}\hfill
  \textbf{\pageref{#3}}\par
}

\newcommand{\suppsubsectionentry}[3]{%
  \noindent\hspace*{1.5em}%
  \makebox[3.8em][l]{%
    \hyperref[#3]{\textcolor{suppred}{\bfseries #1}}%
  }%
  \hyperref[#3]{\textcolor{suppred}{#2}}%
  \nobreak\leaders\hbox to 0.45em{\hss.\hss}\hfill
  \pageref{#3}\par
}


\setcounter{section}{0}
\setcounter{subsection}{0}
\setcounter{subsubsection}{0}
\setcounter{figure}{0}
\setcounter{table}{0}
\setcounter{equation}{0}

\renewcommand{\thesection}{\Alph{section}}
\renewcommand{\thesubsection}{%
  \thesection.\arabic{subsection}%
}
\renewcommand{\thesubsubsection}{%
  \thesubsection.\arabic{subsubsection}%
}

\renewcommand{\thefigure}{S\arabic{figure}}
\renewcommand{\thetable}{S\arabic{table}}
\renewcommand{\theequation}{S\arabic{equation}}


\makeatletter

\@ifundefined{theHsection}{}{%
  \renewcommand{\theHsection}{supp.\Alph{section}}%
}

\@ifundefined{theHsubsection}{}{%
  \renewcommand{\theHsubsection}{%
    supp.\Alph{section}.\arabic{subsection}%
  }%
}

\@ifundefined{theHsubsubsection}{}{%
  \renewcommand{\theHsubsubsection}{%
    supp.\Alph{section}.\arabic{subsection}.\arabic{subsubsection}%
  }%
}

\@ifundefined{theHfigure}{}{%
  \renewcommand{\theHfigure}{supp.\arabic{figure}}%
}

\@ifundefined{theHtable}{}{%
  \renewcommand{\theHtable}{supp.\arabic{table}}%
}

\@ifundefined{theHequation}{}{%
  \renewcommand{\theHequation}{supp.\arabic{equation}}%
}

\@ifundefined{c@algorithm}{}{%
  \edef\suppSavedAlgorithm{\number\value{algorithm}}%
  \setcounter{algorithm}{0}%
  \renewcommand{\thealgorithm}{S\arabic{algorithm}}%
  \@ifundefined{theHalgorithm}{}{%
    \renewcommand{\theHalgorithm}{supp.\arabic{algorithm}}%
  }%
}

\makeatother


\begin{center}
{\huge\bfseries Supplementary Materials\par}
\end{center}

\vspace{1.2em}


\begin{center}
\begin{minipage}{0.88\linewidth}

\large

\centering
{\LARGE\bfseries Contents\par}
\vspace{0.45em}
\rule{\linewidth}{0.6pt}\par
\vspace{0.9em}

\raggedright

\raggedright

\suppsectionentry{A}{Overview}{sec:supp_overview}

\vspace{0.55em}

\suppsectionentry{B}{Datasets}{sec:datasets}

\vspace{0.55em}

\suppsectionentry{C}{Multimodal Search Tools}{sec:search_tools}

\vspace{0.55em}

\suppsectionentry{D}{Experimental Setup}{sec:setup}
\vspace{0.25em}
\suppsubsectionentry{D.1}{Base Model}{subsec:base_model}
\suppsubsectionentry{D.2}{Tool Interaction}{tool_interaction}
\suppsubsectionentry{D.3}{Training Configuration}{training_config}
\suppsubsectionentry{D.4}{Evaluation Protocol}{subsec:eval_protocol}
\suppsubsectionentry{D.5}{Test-Time Scaling}{subsec:test_time_scaling}

\vspace{0.55em}

\suppsectionentry{E}{Tool-Use Analysis}{sec:tool_analysis}
\vspace{0.25em}
\suppsubsectionentry{E.1}{Search Statistics}{subsec:search_statistics}
\suppsubsectionentry{E.2}{Verification and Filtering}{subsec:verification_filtering}

\vspace{0.55em}

\suppsectionentry{F}{Additional Results}{sec:addl_results}

\vspace{0.55em}

\suppsectionentry{G}{Prompts}{sec:prompts}

\vspace{0.55em}

\suppsectionentry{H}{Limitations}{sec:limitations}

\vspace{0.55em}
\noindent\rule{\linewidth}{0.5pt}

\end{minipage}
\end{center}

\clearpage


\section{Overview}
\label{sec:supp_overview}

This supplementary document provides additional details on the datasets, search tools, experimental setup and prompts used in the main paper. It also includes further analyses of judge-based evaluation and evidence verification, together with supplementary results on Qwen3-VL-8B and GPT-5.

\section{Datasets}
\label{sec:datasets}

Following Section~4 in the paper, we finetune Qwen2.5-VL-7B-Instruct on the FVQA training split released by MMSearch-R1 and evaluate on four additional benchmarks: InfoSeek, MMSearch, SimpleVQA, and LiveVQA \cite{mmsearch_r1,infoseek,mmsearch,simplevqa,livevqa}. These datasets are chosen to test complementary aspects of search-augmented multimodal question answering, including search calibration, evidence use, and transfer beyond the training set.

\paragraph{\textbf{FVQA (Finetuning)}.}
We use FVQA as the primary RL finetuning set \cite{mmsearch_r1}. It contains a useful mix of search-free and search-required examples, which is important for learning when search is needed rather than over-invoking tools. In our setup, FVQA also supports the search-aware reward in Section~3.2 of the paper: we perform an initial forward pass without tool use and use the resulting signal as a proxy label when shaping search decisions.

\paragraph{\textbf{InfoSeek}.}
We evaluate on InfoSeek as an additional in-distribution benchmark following the convention of MMSearch-R1 \cite{infoseek,mmsearch_r1}. FVQA is closely tied to the InfoSeek benchmark family, while InfoSeek itself remains broader in scope, making it a useful test of whether the learned behavior transfers beyond the specific FVQA training subset.

\paragraph{\textbf{MMSearch}.}
We evaluate on MMSearch as an out-of-distribution benchmark for multimodal web search \cite{mmsearch}. Compared with FVQA and InfoSeek, MMSearch places greater emphasis on a mix of text only questions without images and multimodal questions, noisy evidence, and effective integration of search results. 

\paragraph{\textbf{SimpleVQA}.}
We additionally evaluate on SimpleVQA as an out-of-distribution benchmark that broadens the evaluation beyond standard image-grounded factual QA \cite{simplevqa}. It contains both image-grounded and text-centric questions, includes multiple-choice examples, and covers both English and Chinese, making it a useful test of transfer across question formats, modality balance, and language.

\paragraph{\textbf{LiveVQA}.}
We include LiveVQA as a recent out-of-distribution benchmark with stronger dependence on timely external knowledge \cite{livevqa}. This makes it a useful stress test of whether the learned search behavior remains effective when up-to-date retrieval matters more directly.

\paragraph{\textbf{Benchmark scope}.}
Taken together, these benchmarks provide a broad test bed for search-augmented multimodal question answering. They probe whether a model can combine image understanding with external retrieval, adapt its search behavior across different question types, and use retrieved evidence effectively to support the final answer. At the same time, the current benchmark suite remains centered on factual question answering with search and therefore does not fully cover the broader space of multimodal agent behavior, such as longer-horizon interaction, richer tool use, or persistent memory across steps. We view extending evaluation in these directions as an important avenue for future work.

\section{Multimodal Search Tools}
\label{sec:search_tools}

\paragraph{\textbf{Search tools}.}
Our agent is equipped with two external tools: image search and text search. Image search is implemented with SerpAPI and returns visually relevant web results for the input image. Text search follows a search-parse-summarize pipeline. At each step, the model may answer directly, invoke image search, or generate a text query $q$ for textual retrieval. When a text query is issued, SerpAPI returns a ranked list of webpages, Jina Reader converts the retrieved pages into LLM-readable markdown, and a Qwen3-32B model summarizes the content into short natural-language tool observations, typically around 300--500 words~\cite{yang2025qwen3technicalreport}. The summarization model is served with vLLM on 8 x A100 GPUs.

\paragraph{\textbf{Caching and efficiency}.}
To reduce latency during RL training, we cache image search results for all FVQA training samples at the start of training. Text search is the main systems bottleneck in both wall-clock time and API cost. A typical SVRL run with 400 training steps, batch size 32, and 4 GRPO rollouts per example can trigger roughly 10,000 to 25,000 text-search calls, depending on the learned search frequency. To reduce this overhead, we cache SerpAPI retrieval results in DynamoDB using the generated text query as the primary key and the original question as a secondary key. When an exact query match is unavailable, we perform approximate matching and reuse cached results if the Jaccard similarity exceeds 0.85. In practice, this reduces text-search cost by up to 40\%.

\paragraph{\textbf{Returned evidence and noise}.}
For each text search, we typically extract up to six webpage summaries and randomly sample between two and six summaries to return to the model. We found this preferable to always returning a fixed number of summaries, since a deterministic interface can encourage overfitting to a narrow evidence pattern. Retrieved summaries are also often noisy: some webpages are not robot-readable, some contain cookie or login interstitials, and some are only weakly related to the generated query. As a result, the returned tool outputs can be partially relevant or entirely unhelpful, which makes evidence filtering and verification central to the setting studied in Section~3 of the paper.

\section{Experimental Setup}
\label{sec:setup}

In this section, we describe the base model, prompting format, training configuration, and evaluation protocol used in our experiments.

\subsection{Base model}
\label{subsec:base_model}
Our primary experiments use Qwen2.5-VL-7B-Instruct as the base multimodal backbone, which we finetune with SVRL and evaluate across all benchmarks \cite{qwen25vl,mmsearch_r1}. Following the setup in Section 4 of the paper, the model operates over short multi-turn trajectories with a maximum of three interaction steps. In each episode, the model may answer directly, invoke image search once, or invoke text search up to two times before termination. We use a maximum context budget of 8124 tokens. In practice, this limited horizon is partly a systems choice, since longer contexts reduce rollout throughput and the effective batch size on a single 8 x A100 node.

\subsection{Tool Interaction}
\label{tool_interaction}
We use a ReAct-style prompting format to structure intermediate reasoning and tool use \cite{react,mmsearch_r1}. Each tool call must be preceded by a reasoning block written as \texttt{<reason> ... </reason>}. The model may then emit one of the following actions:
\begin{enumerate}
\item \texttt{<search><img></search>} to invoke image search,
\item \texttt{<text\_search> q1, ..., q5 </text\_search>} to invoke text search
\item \texttt{<answer> ... </answer>} to terminate with a final answer.
\end{enumerate}
After any tool call, the next turn must also output a binary verification mask of the form \texttt{<verify> 0,1,0,... </verify>} before the next action or final answer. During training, the model proposes five candidate text queries and one valid query is randomly sampled and executed. During standard inference, we use the same protocol. In the test-time scaling setting discussed later, all candidate queries may be executed and aggregated by majority vote. A representative trajectory without detailed contents is shown in Table~\ref{tab:example_trajectory}. See also Figure~1 of the paper for a full example.

\begin{table}[!t]
\caption{\textbf{Representative multi-turn trajectory format used by the model.}}
\label{tab:example_trajectory}
\centering
\small
\setlength{\tabcolsep}{6pt}
\renewcommand{\arraystretch}{1.15}
\begin{tabular}{|c|p{0.72\linewidth}|}
\hline
\textbf{Round} & \textbf{Model output} \\
\hline
Round 1 &
\texttt{<reason> ... need to perform image search. </reason>} \newline
\texttt{<search><img></search>} \\
\hline
Round 2 &
\texttt{<reason> Sources 1 and 3 are helpful ... need to perform text search. </reason>} \newline
\texttt{<verify> 1,0,1,0 </verify>} \newline
\texttt{<text\_search> q1, q2, q3, q4, q5 </text\_search>} \\
\hline
Final round &
\texttt{<reason> Sources 3 and 4 present ... </reason>} \newline
\texttt{<verify> 0,0,1,1 </verify>} \newline
\texttt{<answer> 18-19 days </answer>} \\
\hline
\end{tabular}
\end{table}

\subsection{Training Configuration}
\label{training_config}
All SVRL finetuning experiments are run on a single node with 8 x A100 GPUs. Unless otherwise noted, the main runs use batch size 32, GRPO group size 4, and 400 optimization steps. We keep KL control enabled and use exact-match correctness during training rather than an LLM judge, as discussed in Section 3 of the paper. The remaining optimization settings and reward coefficients are summarized in Table~\ref{tab:svrl_hyperparams}.

The final trajectory reward follows Section 3 of the paper and combines exact-match accuracy, a format reward, and the multiplicative SVRL shaping terms:
\[
r(\tau) = (1-\alpha) r_{\mathrm{acc}}(\tau) r_{\mathrm{svrl}}(\tau) + \alpha r_{\mathrm{fmt}}(\tau),
\]
where
\[
r_{\mathrm{svrl}} = r_{\mathrm{aware}} \cdot r_{\mathrm{count}} \cdot r_{\mathrm{qalign}} \cdot r_{\mathrm{salign}}.
\]

\begin{table}[!t]
\caption{\textbf{Main reward terms and hyperparameters used in SVRL finetuning.} Reward notation follows Section 3 of the paper.}
\label{tab:svrl_hyperparams}
\centering
\small
\setlength{\tabcolsep}{6pt}
\renewcommand{\arraystretch}{1.08}
\begin{tabular}{lll}
\hline
Item & Symbol & Value \\
\hline

Exact-match reward & $r_{\mathrm{acc}}$ & $\{0,1\}$ \\
Search-aware factor & $r_{\mathrm{aware}}$ & $\lambda_{\mathrm{sa}}$ or $1$ \\
Search-aware coefficient & $\lambda_{\mathrm{sa}}$ & $0.1$ \\
Query-count factor & $r_{\mathrm{count}}$ & $1$ or $\max(\epsilon,\hat{k}/k)$ \\
Query-count range &  & $[\epsilon, 1]$ \\
Query-alignment factor & $r_{\mathrm{qalign}}$ & $[\epsilon,1]$ \\
Snippet-alignment factor & $r_{\mathrm{salign}}$ & $[\epsilon,1]$ \\
Format reward & $r_{\mathrm{fmt}}$ & $[0,1]$ \\
Format weight & $\alpha$ & $0.1$ \\
\hline
Training steps &  & 400 \\
Global batch size & $B$ & 32 \\
Dr.GRPO group size & $G$ & 4 \\
Maximum context budget &  & 8124 \\
Maximum prompt length &  & 4096 \\
Maximum response length &  & 2048 \\
Actor learning rate & $\eta$ & 2e-6 \\
Decay ratio &  & 0.95 \\
Warmup steps &  & 45 \\
Mini-batch size &  & 32 \\
Micro-batch size per GPU &  & 8 \\
KL coefficient & $\beta$ & 0.001 \\
Entropy coefficient &  & 0.0 \\
Maximum interaction steps & $T$ & 3 \\
Image search limit &  & 1 \\
Text search limit &  & 2 \\
Candidate text queries per call & $k$ & 5 \\
Training reward &  & Exact match \\
Evaluation judge &  & GPT-5.0 \\

\hline
\end{tabular}
\end{table}

\subsection{Evaluation protocol}
\label{subsec:eval_protocol}

During evaluation, final answers are scored with GPT-5.0 using a fixed system prompt, following the judge-based evaluation protocol adopted by MMSearch-R1 \cite{mmsearch_r1}. We also report exact-match (EM) accuracy as a stricter lexical metric. In our setting, GPT-5.0 judge-based evaluation is used as the primary reported metric, while EM serves as a complementary measure under stricter matching. All other evaluation settings follow the benchmark protocols described in the paper.

\subsection{Test-time scaling details}
\label{subsec:test_time_scaling}

We perform preliminary test-time scaling experiments on a 256-example subset of FVQA-test by increasing the text-search budget at inference time. For a given budget $K \in \{1,\ldots,15\}$, we allow up to $K$ parallel text searches for each question. SVRL already proposes up to five candidate text queries within a single reasoning step, so increasing $K$ first uses these candidate queries directly and then generates additional queries through repeated higher-temperature rollouts. Specifically, as the allowed number of searches increases, we sample additional search proposals with temperatures gradually increased from 0 to 1. Each query is then executed independently through the same retrieval pipeline, producing webpage summaries that are rolled out in parallel until a final answer is obtained. This results in between 1 and 15 candidate answers per question, depending on the allowed search budget. Since each branch triggers additional retrieval and summarization calls, inference cost increases with the search budget.

For aggregation, in practice, we first check whether any answer appears multiple times under exact string matching, and if so, we return the most frequent answer. If no exact majority exists, we consider a word embedding for each candidate answer and use a nearest-neighbor style consensus rule over the candidate set. Concretely, for each answer embedding, we identify the closest other answer embeddings in the set and count how often each candidate lies near the rest. The final prediction is the answer that is closest to the largest number of other candidate answers. We used ChatGPT's text-embedding-3-small in our experiments. Although simple, this procedure is effective for revealing test-time scaling behavior while remaining lightweight compared with more structured aggregation rules. We leave stronger aggregation methods and explicit filtering of incorrect retrieved snippets for future work.

\section{Tool-Use Analysis}
\label{sec:tool_analysis}

In this section, we provide additional analysis of search and evidence-selection behavior, complementing the analysis results in Section 4 of the paper and Figure~5 of the paper.

\subsection{Search Statistics}
\label{subsec:search_statistics}

Figure~5 of the paper summarizes search calibration using search precision and search recall. Search recall measures how often the model invokes search on examples for which search is required, while search precision measures how often a search invocation occurs on an example that truly benefits from search. In this sense, recall captures whether the model searches when needed, whereas precision captures whether it avoids unnecessary search. The plots in Figure~5 aggregate results from four experiments in total, consisting of two FVQA runs and two InfoSeek runs. As discussed in Section~3 of the paper, the search-required labels are based on the search-aware proxy used during training, so these statistics should be interpreted as a behavioral analysis of the learned policy rather than as an absolute measure of search necessity.

We complement this analysis with query-diversity statistics at inference time. On FVQA, the model generates an average of 4.86 unique queries per text-search step. This shows that the policy does not collapse to a single repeated query template despite the fixed five-query interface, and supports the intended role of the query-count reward described in Section~3 of the paper.

\begin{table}[!t]
\caption{\textbf{Search versus no-search behavior on FVQA.} We report the fraction of examples on which the model invokes search, together with accuracy on the search and no-search subsets separately.}
\label{tab:search_vs_nosearch}
\centering
\small
\setlength{\tabcolsep}{6pt}
\renewcommand{\arraystretch}{1.08}
\begin{tabular}{lcc}
\toprule
Metric & MMSearch-R1++ & SVRL \\
\midrule
Used search & 1445 (80.3\%) & 1105 (61.4\%) \\
Accuracy on search subset & 51.0\% & 56.4\% \\
No search & 355 (19.7\%) & 695 (38.6\%) \\
Accuracy on no-search subset & 78.6\% & 79.4\% \\
Overall accuracy & 56.4\% & 65.3\% \\
\bottomrule
\end{tabular}
\end{table}

Table~\ref{tab:search_vs_nosearch} further separates FVQA performance into search and no-search subsets. Compared with MMSearch-R1++, SVRL invokes search much less often (61.4\% versus 80.3\%) while improving accuracy on both subsets, from 51.0\% to 56.4\% on the search subset and from 78.6\% to 79.4\% on the no-search subset. This indicates that the overall gain comes from better search calibration together with stronger performance when search is used, rather than simply increasing tool use.

\begin{table}[!t]
\caption{\textbf{Conditioned query analysis.} Error rates after conditioning on query quality and retrieval success. Lower is better. Query quality and retrieval usefulness are annotated by GPT-5.0 with access to the ground-truth answer.}
\label{tab:query_analysis}
\centering
\scriptsize
\setlength{\tabcolsep}{4pt}
\renewcommand{\arraystretch}{1.06}
\begin{tabular}{p{0.48\linewidth}cc}
\toprule
Condition & MMSearch-R1++ & SVRL \\
\midrule
Good query $\rightarrow$ wrong answer & 38.4\% & 33.7\% \\
Good query + useful result $\rightarrow$ wrong answer & 33.8\% & 29.1\% \\
Good query $\rightarrow$ no useful result & 8.3\% & 6.9\% \\
\bottomrule
\end{tabular}
\end{table}

Finally, Table~\ref{tab:query_analysis} reports a conditioned query analysis in which query quality and retrieval usefulness are annotated by GPT-5.0 with access to the ground-truth answer. Even after conditioning on cases where the issued query is judged to be good, SVRL still improves over MMSearch-R1++. In particular, it reduces answer error both when conditioning only on a good query and when further conditioning on retrieval of at least one useful result. It also slightly reduces the rate at which a good query fails to retrieve any useful result. Taken together, these results suggest that the gains from SVRL are not limited to better query generation, but also extend to stronger downstream evidence use.

\subsection{Verification and Filtering}
\label{subsec:verification_filtering}

We finally analyze the verification masks emitted after each tool call by comparing the model's binary \texttt{<verify>} outputs against GPT-5.0 filtering labels over the same retrieved summaries. Table~\ref{tab:filter_alignment} summarizes the results over all summaries jointly scored by both the model and GPT-5.0. Overall agreement is 71.3\%, which indicates substantial alignment with the teacher signal used during training. However, the disagreements are strongly asymmetric. The model rarely over-filters, with only 82 cases where it rejects a summary that GPT-5.0 considers useful, but it under-filters much more often, with 1495 cases where it accepts a summary that GPT-5.0 labels as noise. This pattern suggests that the learned filter is conservative in discarding evidence and tends to retain questionable summaries rather than risk suppressing potentially useful information. We return to this asymmetry in the Limitations section below.

\begin{table}[!t]
\caption{Alignment between SVRL's verification mask and GPT-5.0 oracle filtering labels over retrieved summaries. The main disagreement is under-filtering: the SVRL finetuned model is much more likely to retain noisy evidence than to discard useful evidence.}
\label{tab:filter_alignment}
\centering
\normalsize
\setlength{\tabcolsep}{6pt}
\renewcommand{\arraystretch}{1.08}
\begin{tabular}{lc}
\hline
Metric & Value \\
\hline
Total snippets scored by both & 5490 \\
Agreement b/w SVRL and GPT-5.0 & 3913/5490 (71.3\%) \\
Both relevant (model = 1, GPT-5.0 = 1) & 2851 \\
Both noise (model = 0, GPT-5.0 = 0) & 1062 \\
Under-filtered (model = 1, GPT-5.0 = 0) & 1495 \\
Over-filtered (model = 0, GPT-5.0 = 1) & 82 \\
\hline
\end{tabular}
\end{table}

\section{Additional Results}
\label{sec:addl_results}

In this section, we provide additional quantitative and qualitative results that complement those in the paper.

\paragraph{\textbf{SVRL on Qwen3-VL-8B}.}
We also finetune Qwen3-VL-8B using the same SVRL training procedure as in the main experiments~\cite{bai2025qwen3vltechnicalreport}. The results are reported in Table~\ref{tab:judge_short}. Overall, SVRL-8B performs very similarly to SVRL-7B and substantially outperforms MMSearch-R1++ on both FVQA and InfoSeek under GPT-5.0 judge-based evaluation. In particular, SVRL-8B reaches 64.9 on FVQA and 64.0 on InfoSeek, compared with 56.4 and 55.1 for MMSearch-R1++. These results suggest that the gains from SVRL are not tied to a single backbone and transfer to a different multimodal base model under the same finetuning setup.

\paragraph{\textbf{Comparison to GPT-5.2}.}
Table~\ref{tab:judge_short} also reports inference-only results for GPT-5.2 and GPT-5.2-thinking as stronger proprietary reference models. A direct-answer GPT-5.2 baseline improves over direct-answer Qwen3-VL-8B, but remains far below search-based methods. Under adaptive search, GPT-5.2-thinking performs competitively, but both SVRL-7B and SVRL-8B achieve higher GPT-5.0 judge-based accuracy on FVQA and InfoSeek. This comparison is notable because it shows that the proposed training procedure can make compact open-source models competitive with, and in these settings stronger than, a much larger proprietary system when both are evaluated in a search-based regime.

\begin{table}[t]
\caption{\textbf{Accuracy (\%)} on FVQA and InfoSeek under GPT-5.0 judge-based evaluation and exact match (EM). We evaluate against the latest GPT-5.2 thinking and no-thinking modes. Additionally we report accuracies on SVRL finetuned Qwen-3-VL-8B which shows similar results to SVRL-7B presented in the main paper.}
\label{tab:judge_short}
\centering
\normalsize
\setlength{\tabcolsep}{5pt}
\begin{tabular}{l|cc|cc}
\toprule
Model & \multicolumn{2}{c|}{FVQA} & \multicolumn{2}{c}{InfoSeek} \\
 & GPT-5.0 & EM & GPT-5.0 & EM \\
\midrule

\multicolumn{5}{c}{\textbf{Direct Answer}} \\
\midrule
Qwen3-VL-8B & 22.2 & 15.4 & 22.2 & 13.9 \\
GPT-5.2 & 34.0 & 14.3 & 29.9 & 12.4 \\

\midrule
\multicolumn{5}{c}{\textbf{Adaptive Search}} \\
\midrule
MMSearch-R1++ & 56.4 & 40.6 & 55.1 & 30.4 \\
\textbf{SVRL-full-7B} & \textbf{65.3} & 43.0 & \textbf{64.7} & 41.7 \\
\textbf{SVRL-full-8B} & \textbf{64.9} & 40.2 & \textbf{64.0} & 36.5 \\
GPT-5.2-thinking & 60.8 & 21.2 & 52.2 & 15.8 \\

\bottomrule
\end{tabular}
\end{table}

\paragraph{\textbf{Judge-based evaluation and exact match}.}
The table also reinforces the distinction between judge-based evaluation and exact match. Across nearly all methods, exact-match accuracy is substantially lower than GPT-5.0 judge-based accuracy, especially for proprietary models. This gap is consistent with the discussion in the evaluation section: exact match is sensitive to surface-form variation, while the judge-based metric is more tolerant to paraphrases and semantically equivalent answers. Importantly, the overall ranking of the stronger search-based methods remains favorable to SVRL under both evaluation protocols.

\section{Prompts}
\label{sec:prompts}

We follow the same overall prompt structure as MMSearch-R1 for training, inference, and judge-based evaluation \cite{mmsearch_r1}, with minor changes to match the ReAct-style interaction format used in our method. Table~\ref{tab:svrl_prompt_main} shows the main prompt used for search-based training and inference. For the direct-answer baseline and judge-based evaluation, we use the same prompt templates as in MMSearch-R1, with GPT-5.0 replacing the original judge model where applicable. Table~\ref{tab:svrl_prompt_main} summarizes the system prompt used for search-based training and inference in SVRL.

\begin{table}[!t]
\caption{Main system prompt used for search-based training and inference in SVRL.}
\label{tab:svrl_prompt_main}
\centering
\scriptsize
\setlength{\tabcolsep}{4pt}
\renewcommand{\arraystretch}{1.25}
\begin{tabular}{|p{0.94\linewidth}|}
\hline
\textbf{System message.} You are a helpful assistant who will be evaluated on a Visual Question Answering Task. You should strictly follow a reason-to-act process to answer a user-provided question about an image. All thinking must be written inside \texttt{<reason>} and \texttt{</reason>} tags. \\[0.6ex]
\hline
\textbf{Available actions.} First analyze the question and any available observations, including the user-provided image and any retrieved search results. Then choose one of the following actions: \texttt{<search><img></search>} to perform image search; \texttt{<text\_search>query1, query2, query3, query4, query5</text\_search>} to generate five unique relevant text queries, one of which is selected at random during training; or \texttt{<answer>xxxxx</answer>} to return the final answer. If an action was performed previously, output a binary relevance list containing as many comma separated values as the number of tool output snippets like this: \texttt{<verify>0/1, 0/1, ...</verify>} before choosing the next action. Retrieved image and text results are placed inside \texttt{<information>} and \texttt{</information>} tags. \\[0.6ex]
\hline
\textbf{Allowed output formats.} The model output must follow exactly one of the following formats: \newline
\texttt{<reason> YOUR THINKING PROCESS </reason>} \newline
\texttt{<search><img></search>} \newline
or \newline
\texttt{<reason> YOUR THINKING PROCESS </reason>} \newline
\texttt{<verify>0/1, 0/1, ...</verify>} \newline
\texttt{<text\_search> FIVE GENERATED COMMA SEPARATED UNIQUE TEXT QUERIES </text\_search>} \newline
or \newline
\texttt{<reason> YOUR THINKING PROCESS TO ASSESS TOOL OUTPUTS AND ANSWER THE QUESTION </reason>} \newline
\texttt{<verify>0/1, 0/1, ...</verify>} \newline
\texttt{<answer> YOUR ANSWER AFTER GETTING ENOUGH INFORMATION </answer>} \\[0.6ex]
\hline
\textbf{Additional instructions.} If an image or text search call was made previously, the model should first reason about why each returned summary was helpful or unhelpful and then output a binary relevance list inside \texttt{<verify>} and \texttt{</verify>} before producing the final answer. If no prior search call was made, output an empty \texttt{<verify>} \texttt{</verify>} block. The number of binary scores must match the number of retrieved summaries. The final answer must appear only inside \texttt{<answer>} and \texttt{</answer>} tags, without extra explanation. If the question expects a year, date, quantity or location, be as specific as possible and mention appropriate units. If the question is yes-or-no, answer only yes or no. \\[0.6ex]
\hline
\end{tabular}
\end{table}

\section{Limitations}
\label{sec:limitations}

Our method still has several limitations. First, although verification is performed by the agent itself at inference time, the query- and snippet-alignment rewards during training rely on verifier-generated labels with access to the ground-truth answer. This improves reward quality, but also introduces a dependence on the teacher model and its annotation biases. Similarly, our primary evaluation metric uses GPT-5.0 as a judge, which is more robust than exact matching for aliases and paraphrases but is not fully reproducible and can be sensitive to prompt phrasing. Future work could reduce this dependence by developing stronger self-supervised or process-level verification signals, lightweight learned verifiers, or label-free process rewards \cite{training_verifiers,lets_verify,freeprocreward_chen2024}. Second, the retrieval setting remains noisy and non-stationary: web pages can be irrelevant, inaccessible, outdated, or poorly parsed, and our appendix analysis shows that the learned filter is still more likely to retain noisy evidence than to discard useful evidence. Improving robustness under retrieval shift may require tighter integration of retrieval, reranking, and verification, as well as training on richer web-agent settings with more realistic search dynamics \cite{avis,mmsearch_r1,deepmmsearch_r1,webwatcher}.

A second limitation is scope. Our experiments focus on search-augmented multimodal question answering with a short interaction horizon and a restricted action space centered on image search, text search, verification, and answer generation. As a result, the current results should be interpreted as evidence of improved search-based multimodal reasoning rather than a complete treatment of general multimodal agency. In addition, our test-time scaling experiments use a simple consensus-based aggregation rule over candidate answers. While this is sufficient to show that extra search budget can improve accuracy, more structured search and aggregation procedures may further strengthen performance \cite{self_consistency,tot,segb_xie2023}. Extending SVRL to richer tool suites, longer-horizon interaction, and stronger inference-time aggregation remains an important direction for future work \cite{toolformer,avis,webwatcher}.


\clearpage

\setcounter{section}{\suppSavedSection}
\setcounter{subsection}{\suppSavedSubsection}
\setcounter{subsubsection}{\suppSavedSubsubsection}
\setcounter{figure}{\suppSavedFigure}
\setcounter{table}{\suppSavedTable}
\setcounter{equation}{\suppSavedEquation}

\makeatletter
\@ifundefined{c@algorithm}{}{%
  \setcounter{algorithm}{\suppSavedAlgorithm}%
}
\makeatother

\endgroup
